# Synthetic Defect Generation for Display Front-of-Screen Quality Inspection: A Survey

*Shancong Mou\*, Meng Cao\*\*, Zhendong Hong\*\*, Ping Huang\*\*, Jiulong Shan\*\* and Jianjun Shi\**

*\*Georgia Institute of Technology, Atlanta GA*; *\*\*Apple, Cupertino, CA*

## Abstract

*Display front-of-screen (FOS) quality inspection is essential for the mass production of displays in the manufacturing process. However, the severe imbalanced data, especially the limited number of defect samples, has been a long-standing problem that hinders the successful application of deep learning algorithms. Synthetic defect data generation can help address this issue. This paper reviews the state-of-the-art synthetic data generation methods and the evaluation metrics that can potentially be applied to display FOS quality inspection tasks.*

## Author Keywords

Display quality inspection, synthetic defect generation review.

## 1. Introduction

The display industry has experienced explosive growth in the past few decades, spurred by increasing demand in new consumer and industrial applications. The backbone of this paradigm shift is a continuous evolution to introduce entirely new technologies (organic light-emitting diode (OLED), microLED, etc.) or significant enhancements to existing technologies (miniLED, Quantum Dots, etc.) with improved quality at a decreasing cost. All of these factors pose challenges for quality control of the display manufacturing process, where image-based inspection is an essential task [6, 7]. To achieve accurate defect detection for display panels, deep learning can be used [8]. Even though there are considerable advancements in unsupervised/semi-supervised learning approaches in anomaly/defect detection, supervised learning-based defect detection techniques are more popular due to their superior performance in both accuracy and robustness, given enough training samples. However, the limited number of defective samples has been a long-standing problem that hinders the successful application of modern machine learning algorithms in display inspection [1]. It takes a long time to accumulate enough defective samples to train an effective model, which can cause production delays. Therefore, synthetic defect generation is a valuable approach.

Synthetic defect generation can be categorized into rule-based and data-driven methods. Traditional rule-based synthetic defect generation methods use rules to guide the generation of artificial defects. Those rules usually come from industrial use cases and are extracted by experienced engineers. For example, Jo et al. [6] generated five different types of display defects (vivid dot, faint dot, line, stain, and mura) by specifying a set of rules that control the shape and appearance of defects on each sample. The generated defective samples are shown in Figure 1.

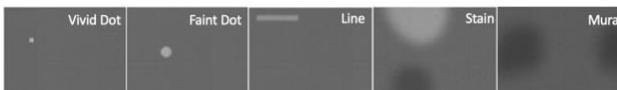

**Figure 1.** Rule-based display defect generation [6]

There are three limitations of the rule-based method: (i) it requires deep expert knowledge about the display manufacturing process; (ii) the generated defects lack randomness to mimic the real defect; and (iii) a slight variation in an existing defect not fully described by the preset rules may likely turn into escapes, resulting in quality degradation. Another approach, data-driven synthetic defect generation methods directly learn from existing data. It will accelerate the development cycle in defect detection applications once implemented. Therefore, data-driven synthetic defect generation methods are popular for data augmentation in defect detection applications where defective samples are rare.

Synthetic defect generation belongs to the category of data augmentation [9]. Different from data augmentation for general purposes, synthetic defect generation focuses on generating vivid defective samples instead of blindly enlarging the whole dataset, which adapts to the data-rich but label (defect) rare display manufacturing environment. Therefore, basic image manipulations [9] such as geometric transformations, color space augmentations, kernel filters, mixing images, random erasing, and feature space augmentation are not applicable, and learning-based methods are more suitable in this case.

In this paper, we provide a review of synthetic defect generation using deep learning with potential applications in display manufacturing. The discussion of synthetic defect generation for display application is rare [1]. Therefore, our review will include but is not limited to display manufacturing. We hope that the synthetic defect generation methods adopted in other fields will inspire the audience from the display manufacturing field. In this review, the application of generative adversarial networks (GANs) [10] and their variants will be heavily covered. The paper will review synthetic defect generation methods (Section 2), introduce evaluation metrics for generated defects (Section 3), and provide a summary (Section 4).

## 2. Review of synthetic defect generation methods

This section will first introduce each synthetic defect generation method and then discuss the corresponding applications using this method for defect generation. The similarities and differences between the display manufacturing application and the specific application scenario will also be discussed.

Generative adversarial networks (GAN) were first proposed by Goodfellow et al. [10] to mimic the real data distribution. Since then, it has become popular in various fields [11]. According to Yann LeCun, 'GANs are the most interesting idea in the last ten years in machine learning'. More recently, GANs have been improved to (i) generate vivid images by using deep convolutional neural networks (DCGAN) [12], (ii) enhance the training stability by using PGAN [13] and Wasserstein GAN (WGAN) [14], (iii) conduct unpaired image-to-image translation by using Cycle-Consistent Adversarial Networks (CycleGAN) [15], and (iv) generate multi-class images by using Conditional

GAN (CGAN) [16] and so on. A comprehensive review of GANs can be found in [11]. This paper will introduce some critical GAN variants and review their applications for synthetic defect generation in manufacturing applications. Table 1 provides a summary of those methods and their applications.

**Table 1.** GAN methods used in synthetic defect generation

| Methods / Applications | GAN/ DCGAN /WGAN | Cycle GAN | CGAN /ACGAN |
|---|---|---|---|
| Display [1] | | | √ |
| Wafer [2,4] | | | √ |
| Assembly and test [5] | | √ | √ |
| Conveyor belt [17] | √ | | |
| Electrical machines [18] | √ | | |
| Solar cells [19] | √ | | |
| Non-destructive testing [20,21] | √ | | |
| Pear [22]/ Coffee [23] | √ | | |
| Railway [24] | | √ | |
| Commutator [25] | | √ | |
| Chiller [26] | √ | | √ |
| Fiber layer up [27] | | | √ |
| Laser welding [28] | | | √ |
| General methods [29, 37] | | √ | √ |

### 2.1. Vanilla GAN and its variants

GANs' structure (DCGAN/WGAN) is shown in Figure 2. It consists of two models: a generator network (G) tries to generate images that look real, and a discriminator network (D) tries to distinguish between the real image and the fake image generated by the generator. The generator and discriminator will compete to achieve a Nash equilibrium [11]. Finally, the generator captures the distribution of input samples.

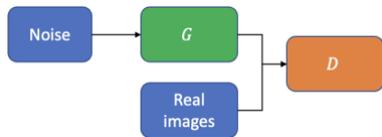

**Figure 2.** Structure of GAN

In terms of utilizing GAN generating defective samples in a manufacturing process, Bo et al. [17] utilized the DCGAN [12] to learn from local defective images to create the defect region mask and fuse with the normal background to produce defective conveyor belt images (Figure 3). By adopting this approach, a single background can generate many different defect samples, which is the desired property in display manufacturing. However, to achieve so, a localization-level (bundling box) labeled defect dataset is needed. This method has the potential to be applied to synthetic defect generation, where a localization level mask can be generated by unsupervised learning [30].

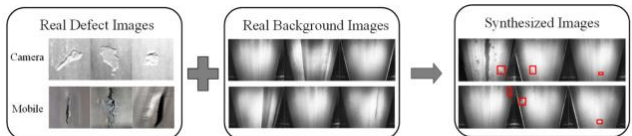

**Figure 3**. Defective conveyor belt images generation [17].

Tang et al. [19] used DCGAN to generate synthetic defective solar cells. The generated images are shown in Figure 4.

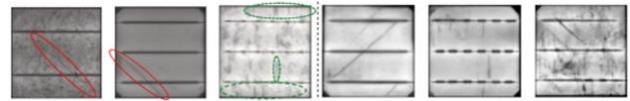

(a) real defects     (b) generated defects
**Figure 4.** Generated defect solar cell images [19].

Vanilla GAN can generate defective samples and augment the dataset. However, its performance may be limited when the real defective images are rare, which is typical in display manufacturing. In this case, directly generating the defective images from noise may not be optimal. An alternative approach is to learn the mapping from normal images to defective images by using CycleGAN.

### 2.2. CycleGAN and its variants

CycleGAN's structure is shown in Figure 5. It has two generators ($G_n$, $G_d$) and two discriminators ($D_n$, $D_d$). $G_d$ tries to transform the real normal images into defective images, and $D_d$ tries to distinguish between real defective images and the fake defective images generated by $G_d$. Similarly, $G_n$ tries to transform real defective images into normal images, and $D_n$ tries to distinguish between real normal images and the fake normal images generated by $G_n$. More importantly, a cycle consistency loss [15] is used to avoid the requirement of a training set with perfectly aligned pairs of defective and normal images. Since CycleGAN performs image translation from normal images to defective ones instead of generating defective images from noise, it requires fewer defective training samples.

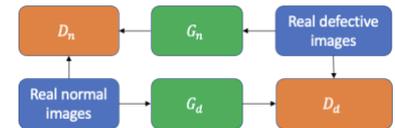

**Figure 5.** Structure of CycleGAN

In utilizing CycleGAN to generate defective samples, Hoshi et al. [24] directly applied CycleGAN to generate synthetic railway scratches defect samples. Singh et al. [5] demonstrated the effectiveness of CycleGAN [15] for synthetic defect generation in assembly and test manufacturing applications when pairs of defect and defect-free images are not available as commonly seen in display manufacturing applications. To control and generate various types of defects (Figure 6), they also integrate the CGAN [16] with CycleGAN. Niu et al. [25] proposed to use an SDGAN (Figure 7) to generate commutator surface defects.

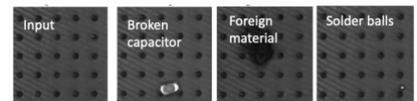

**Figure 6.** Generated synthetic assembly manufacturing defects [5].

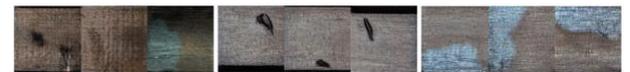

**Figure 7.** Generated commutator surface defects [25]

Then, Liu et al. [31] proposed an FD-Cycle-GAN method to generate railway defective fastener images. This work generates a diverse dataset by adding a diversity loss. It is inspiring for synthetic display defect generation as the acquired display defect dataset usually lacks diversity and has a long-tail distribution.

However, a significant problem with vanilla CycleGAN is that it cannot generate multiple defective images from a single or similar normal image. This problem will be a challenge for utilizing this approach in synthetic display defect generation because the normal images of displays are similar. To solve this issue, Zhang et al. [37] proposed to generate defect foregrounds under the guidance of a spatial and categorical control map and then combine the generated defect foregrounds with normal images to generate defective images. Augmented CycleGAN [32] and multimodal unsupervised image-to-image translation [33] can also be used to address this issue.

### 2.3. CGAN and their variants

CGAN and its variants (InfoGAN [34], ACGAN[35]) are useful when there are multiple types of defects, such as multiple

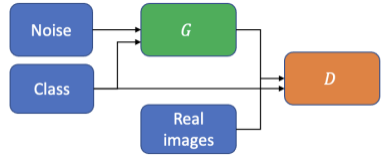

**Figure 8.** Structure of CGAN

types of display defects mentioned in the introduction. The structure of CGAN is shown in Figure 8. It can integrate the class (defect type) information in the defect generation and discrimination process. Meister et al. [27] used conditional

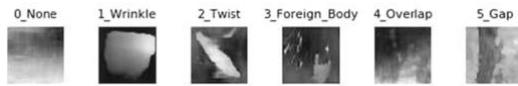

**Figure 9.** Generated defect fiber [27].

DCGAN to generate synthetic defective samples for fiber layup inspection processes. Generated defective images are shown in Figure 9. We can see that different types of defects can be generated, which has the potential to be used for synthetic display defect generation. Xiong et al. [1] proposed a Multi-Modal One Class GAN (MMOCGAN) model to generate defective display samples for data augmentation and shows a significant improvement of the defect detection algorithms. To address the scarce issue of the defective samples in the application, the authors propose to generate synthetic defects images following the complementary distribution of the normal images. However, it is not verified whether the generated defective samples are valid. Liu et al. [4] proposed a focal auxiliary classifier generative adversarial network (FAC-GAN) to generate defective samples in wafer manufacturing and address the imbalance among different defect types. Yu et al. [2] proposed a multi-granularity GAN (MGGAN) to generate wafer map defects. The multi granularity features from the training images are extracted by an auxiliary feature extractor (Resnet101) to stabilize the training and solve the class imbalance problem.

Those methods can be applied in synthetic display defect generation since there are multiple types of defects, and their occurrence frequency may not be the same. A few variants of the CGAN based synthetic data generations and applications are summarized in Table 1.

### 3. Evaluation metrics for generated defects

The quality of generated defective images influences the performance of trained defect detection algorithms. Therefore, how to evaluate the generated defective samples is important. There are *quantitative* and *qualitative* approaches for assessing the quality. For a complete review, please refer to [36].

A common qualitative approach examines the generated images by eye [5, 20, 27]. For example, Singh et al. [5] evaluated the synthetic defect generation algorithms performance by checking the realism of generated samples. This method is straightforward. However, it may be subject to human subjective bias.

Several quantitative metrics that *directly* measure the quality of generated images used in manufacturing applications are listed as follows:

- Fréchet Inception Distance (FID) [2, 18, 19, 25] compares the distance of generated data distribution and real data distribution in a latent space. It has good discriminability, robustness, and consistency with human judgments. The smaller FID is, the better. However, its gaussian assumptions may be too restrictive.
- Maximum Mean Discrepancy (MMD) [19] computes the dissimilarity between two probability distributions of generated data and real data. The lower MMD is, the better.

For example, Tang et al. [19] used FID and MMD to compare the performance of WGAN and DCGAN in generated synthetic defect solar cell images. Sabir et al. [18] used DCGAN to generate a 1-dimensional faulty signal of electrical machines, and Fréchet inception distance (FID) [36] is used to evaluate generated signal quality. Yan [26] combined the CGAN [16] and WGAN [14] to synthesize multiple types of chiller fault samples. To evaluate the quality of the generated samples, the author proposed to train a VAE or a GANomaly model by using the generated defaults and then testing it on real defaults. Those methods evaluate the quality of generated images. However, it does not directly relate to the performance of the defect detection algorithms.

In manufacturing defect detection applications, the goal is to improve the accuracy of defect detection algorithms, which provides natural *indirect* quantitative evaluation metrics. The defect detection algorithm performance (i) with and without synthetic data; (ii) with synthetic data generated from different methods, are usually compared. In defect detection applications, the *indirect* quantitative evaluation metrics can be further categorized into three different levels of accuracy, including instance level (detect/classify the defective samples [2, 4, 19, 22, 24, 25, 27, 28, 31]), localization level (detect the defect location inside each sample [17, 21]), and pixel-wise level (extract the defect segmentation mask [29]). For example, Xiong et al. [1] used the F1 score of the downstream classification model to evaluate the generated synthetic display defect dataset in the instance level. They compare the F1 score of the classification algorithms trained on the GAN augmented dataset with the original dataset/oversampling method augmented dataset and identified a significant performance improvement. Bo et al. [17] compared the average precision, the maximum recall, and the maximum F-score of the defect allocation algorithms among different synthetic generation algorithms at the localization level. Wu et al. [29] used pixel-wise segmentation accuracy such as pixel accuracy (PA) and mean intersection over union (MIoU) to evaluate the proposed synthetic defect generation algorithms. *Indirect* quantitative evaluation is time-consuming compared to *direct* quality measures. However, adopting the *indirect* quantitative evaluation does not add much effort to the process and is widely adopted in manufacturing applications.

### 4. Summary

This paper surveys several GAN-based synthetic defect generation techniques and reviews their applications in the manufacturing field. We also discussed their potential to be applied in synthetic display defect data generation.